\documentclass[10pt,twocolumn,letterpaper]{article}

\usepackage{3dv}
\usepackage{times}
\usepackage{epsfig}
\usepackage{graphicx}
\usepackage{amsmath}
\usepackage{amssymb}

\usepackage[pagebackref=true,breaklinks=true,letterpaper=true,colorlinks,bookmarks=false]{hyperref}

\threedvfinalcopy 

\def\threedvPaperID{164} 

\ifthreedvfinal\fi
\begin{document}

\title{Joint Semantic Segmentation and Depth Estimation with Deep Convolutional Networks}

\author{Arsalan Mousavian\\
George Mason University\\
4400 University Dr, Fairfax, VA 22030\\
{\tt\small amousavi@gmu.edu}
\and
Hamed Pirsiavash\\
University of Maryland Baltimore County\\
1000 Hilltop Cir, Baltimore, MD 21250\\
{\tt\small hpirsiav@umbc.edu}
\and
Jana  Ko\v{s}eck\'a\\
George Mason University\\
4400 University Dr, Fairfax, VA 22030\\
{\tt\small kosecka@gmu.edu}
}

\maketitle

\begin{abstract}
Multi-scale deep CNNs have been used successfully for problems mapping each pixel to a label, such as depth estimation and semantic segmentation. It has also been shown that such architectures are reusable and can be used for multiple tasks. 
These networks are typically trained independently for each task by varying the output layer(s) and training objective. In this work we present a new model for simultaneous depth estimation and semantic segmentation from a single RGB image. Our approach demonstrates the feasibility of training parts of the model for each task and then fine tuning the full, combined model on both tasks simultaneously using a single loss function. Furthermore we couple the deep CNN with fully connected CRF, which captures the contextual relationships and interactions between the semantic and depth cues improving the accuracy of the final results. The proposed model is trained and evaluated on NYUDepth V2 dataset~\cite{SilbermanECCV12} outperforming the state of the art methods on semantic segmentation and achieving comparable results on the task of depth estimation.  
\end{abstract}

\section{Introduction}

Deep convolutional networks (CNNs) attracted a lot of attention in the past few years and have shown significant progress in object categorization enabled by the availability of large scale labeled datasets~\cite{KrizhevskyNIPS12}.  For semantic segmentation problem, which requires learning a pixel-to-pixel mapping,  several approaches have been proposed, 
for handling the loss of resolution and generation of a pixel level labelling~\cite{LongCVPR15,Segnet15}. The initial CNN models for semantic segmentation showed that the response maps in final layers were often not sufficiently well localized for accurate pixel-level segmentation. To achieve more accurate localization property, the final 
layers have been combined with fully connected CRF's~\cite{MurphyICLR15} yielding notable improvements in the segmentation accuracy.  
Independent efforts explored the use of CNNs for depth estimation from a single view~\cite{EigenNIPS14}. 
Most recent work of~\cite{EigenICCV15} showed that common network architecture can be used for problems of semantic segmentation, depth estimation and surface normal estimation. The authors have shown that by changing the output layer and the loss function,  the same network architecture can be trained effectively for different tasks achieving state of the art performance of different benchmark datasets. In contrast, we train the same network under multi task loss for semantic segmentation and depth estimation and our experiments show that multi-task learning boosts the performance.

We follow this line of work further and postulate the simultaneous availability of the depth estimates  can further improve the final labeling. To support that we 
present a new approach and model for simultaneous depth estimation and semantic segmentation from a single RGB image, where the two tasks share the underlying feature representation.  To further overcome the difficulties of deep CNNs to capture the context and respect the low-level segmentation cues as provided by edges and pixel values, we integrate CNN with a fully connected Conditional Random Field (CRF) model and learn its parameters jointly with the network weights. We train the model on NYUDepth V2~\cite{SilbermanECCV12} and evaluate the final quality of both semantic segmentation with estimated depth, without depth and depth estimation alone. The proposed approach outperforms the state of the art semantic segmentation methods \cite{EigenICCV15,LongCVPR15,GuptaECCV14}
and achieves comparable results on the task of depth estimation in~\cite{EigenICCV15}.  

\section{Related work}

In the past few years, convolutional neural networks have been applied to many high level problems in computer vision 
with great success. The initial categorization approaches focused on assigning a single label to an image~\cite{KrizhevskyNIPS12}, followed by application of the same classification strategy to windows or region proposals 
generated by independent segmentation process~\cite{GirshickCVPR14}. In addition to classification problems, these models marked also great success for a variety of regression problems, including pose estimation, stereo, localization 
and instance level segmentation, surface normal segmentation and depth estimation. The initial architectures obtained by concatenating multiple convolutional layers follow by pooling were suitable 
for image classification or regression problems, where single label of vector valued output was sought.
The earlier layers before the fully connected layers were also found effective as feature maps used for variety of other traditional computer vision tasks~\cite{DBLP:journals/corr/AzizpourRSMC14}. 
For the problem of semantic segmentation CNN approaches typically generated features or label predictions at multiple scales~\cite{Couprie2013} and used averaging and superpixels and for obtaining the final boundaries.  In \cite{MostajabiCVPR15} CNNs were applied to superpixels, which were directly classified using feedforward multilayer network. Alternative strategy by~\cite{GuptaECCV14} used  CNN features computed over RGB-D region proposals generated by low-level segmentation methods.  These methods although initially successful relied on the availability of independent segmentation methods to either refine the results or to generate object proposals.

One of the first approaches to tackle the semantic segmentation as a problem of learning a pixel-to-pixel mapping using CNNs was the work of~\cite{LongCVPR15}. There authors proposed to apply 1x1 convolution label classifiers at features maps from different layers and averaging the results.  Another line of approaches to semantic segmentation adopted an auto-encoder style architecture \cite{NohICCV15} \cite{Segnet15} comprised of convolutional and deconvolutional layers.  The deconvolutional part consists of unpooling and deconvolution layers where each unpooling layer is connected to its corresponding pooling layer on the encoding side. The convolutional layers remain identical  to the architectures of \cite{KrizhevskyNIPS12} and \cite{VGGICLR15} and the deconvolutional layers are trained.  Authors in \cite{NohICCV15} formulated the problem of semantic segmentation of the whole image as  collage of individual object proposals, but also use the deconvolution part to delineate the object shape at higher resolution inside of the proposal window. The object proposal hypotheses are then combined by averaging or picking  the maximum to produce the final output. 

The lack of context or the capability of generating more accurate boundaries were some of the typical shortcomings of the above mentioned CNN based semantic segmentation architectures. In the pre-CNN approaches to semantic segmentation Conditional Random Fields (CRF) have been used effectively and provided strong means for integrating the local multi-class predictions with context and local information captured by pixels and edges~\cite{LadickyCVPR14}.  To incorporate the benefits of CRF's for semantic segmentation the authors in  Chen et al \cite{MurphyICLR15} proposed to combine deep CNNs responses of the last convolutional layer with the fully connected CRF. They used the hole method of \cite{GiustiICIP13} to make the VGG network \cite{VGGICLR15} denser and resized the label probability map using bilinear interpolation. The resized semantic probability map was then used in place unary potentials for a fully connected CRF proposed by~\cite{KoltenNips11}. In spite of exhibiting significant improvement over initial results in~\cite{LongCVPR15}, the method of \cite{MurphyICLR15} trained the CNN part and fully connected CRF part independently. Some of the subsequent efforts following this improvement led to joint training of CNNs and CRFs.  Zheng et al. \cite{TorICCV15} addressed this issue by transforming the mean field approximation of \cite{KoltenNips11} to a sequence of differentiable operations which can be incorporated in the CNN training. They  learned with back-propagation the compatibility term of two labels regardless of the cell location. In the follow up work of~\cite{DPNICCV16} authors addressed this shortcoming by learning the compatibility between pairs of labels while considering their relative spatial location. 

Previously reviewed methods for semantic segmentation have been applied to either images or RGB-D images, demonstrating improvements when the depth channel was available~\cite{GuptaECCV14,RenCVPR12}. Separate line of work focused on single image depth estimation. Early works exploited constraints of structured man-made, mostly indoors environments and rich features~\cite{GuptaCVPR15,UrtasunICCV13}. Saxena et al \cite{SaxenaPAMI08} considered general outdoor scenes and formulated the depth estimation as Markov Random Field (MRF) labeling problem, where depth was estimated using a large set handcrafted features computed at multiple scales and hierarchical MRF.   
Attempts to revisit these problems using deep CNNs were considered by Eigen et al \cite{EigenNIPS14}, where depth was estimated using two networks, which handled coarse and fine scale depth estimation. The input to the first network is the whole image and output is a coarse depth map, while the second network, takes the coarse depth map produced by the previous stage and an image patch at 1/4 input image scale to produce the fine details of the depth map. Liu et al \cite{LiuCVPR15} addressed the depth estimation problem as estimating a single floating-point number for each superpixel representing the depth of superpixel center. 
There are few works where both the semantic and depth cues jointly contribute to semantic understanding and scene layout. Zhang et al \cite{FidlerICCV15}, segmented the car instances in an image and provided the depth ordering of each car instance. 
Closest to our work in trying to use both depth and semantic cues are \cite{LadickyCVPR14} and \cite{WangCVPR15}. The authors of \cite{LadickyCVPR14} propose to estimate depth and semantic category using an unbiased semantic depth classifier, whose output on a bounding box remains the same when the image and bounding box scales by $\alpha$. In \cite{WangCVPR15}, a coarse depth map is estimated by a CNN and they add finer depth details by extracting frequent templates for each semantic category. Finding frequent discriminant patches for each category requires more number of images from each category. As a result, their method does not scale well with the increase in number of classes.

The proposed model is the first to estimate the semantic labels and 
depth jointly from a single RGB image using a shared representation. While previous methods coupled CNNs with CRFs and refined the parameters of both components jointly, our approach is the first to do so with a more expressive objective function which incorporates the interactions between the depth and semantic labels. 

\section{Proposed Method} 
Semantic segmentation and depth estimation have been often addressed separately in the past.   In this work, we demonstrate the possibility of training a network for depth estimation and semantic segmentation together, where the two tasks learn a shared underlying feature representation.  
This has a number of  benefits: First of all, a single network handles both tasks which results in reducing the amount of computation and memory footprint by sharing the parameters. Moreover, the performance of semantic segmentation is increased because the network implicitly learns underlying physics by estimating depth for each pixel.

\begin{figure*}[t]
\centering
\includegraphics[width=\textwidth]{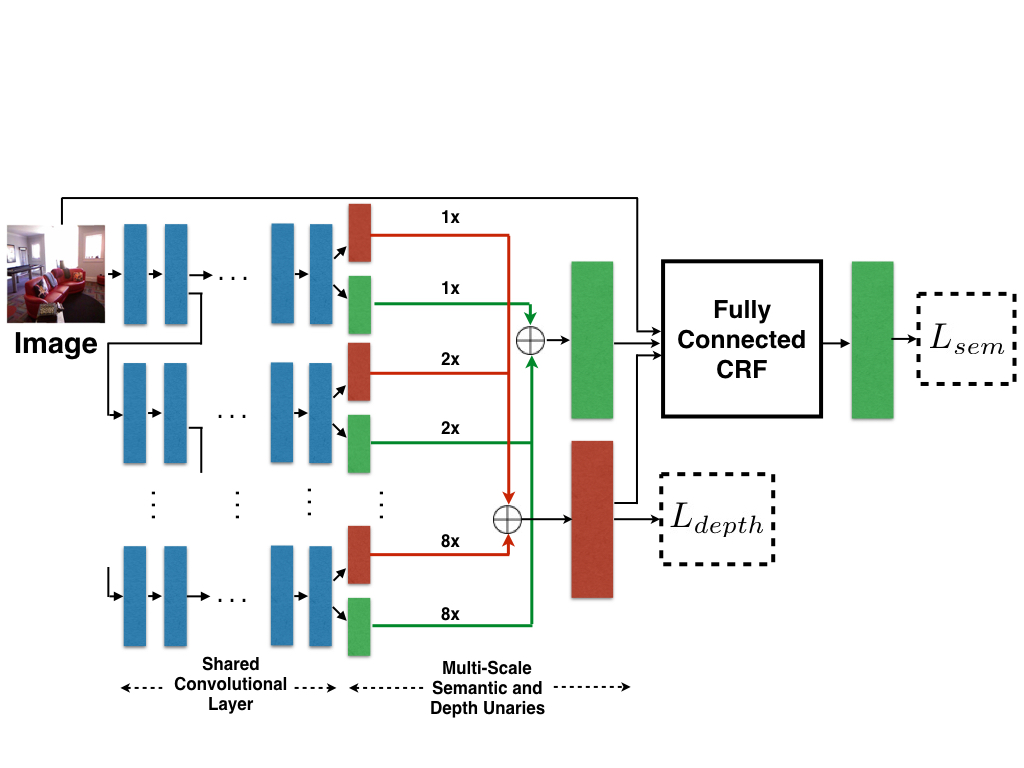}
\caption{Overview of the proposed method. Multi-scale fully convolutional network is used for image representation. The network consists of 5 different paths and each path extracts feature at a difference scale. At the end of each path, two convolutional layers extract feature for semantic segmentation and depth estimation. These feature maps are scaled and aggregated to form a comprehensive feature map for semantic segmentation and depth estimation. Depth values are estimated using Eq. \ref{eq:depth_value}. Estimated depth values along with semantic unaries and the image are passed through the fully connected CRF to get the final probabilities of semantic labels.}
\label{fig:overview}
\end{figure*}

\noindent
The proposed method takes RGB image as an input and uses a single network to make an initial estimate of depth and the semantic label for each pixel. These estimates are then combined to produce a final semantic segmentation. Using the estimated depth helps to resolve confusions between similar semantic categories such as pillow vs sofa, book vs bookshelf, and so on. The parameters of multi-scale network is learned by optimizing a joint objective function for semantic segmentation and depth estimation. The learned weights can be used for both tasks individually or for both, jointly. The proposed approach is an alternative to methods which use the depth channel of RGB-D sensor as an input to the network~\cite{LongCVPR15}. The raw depth channel often provides missing or inaccurate values, which  are replaced by the output of the in-painting algorithms ~\cite{CadenaICRA13}. On the other hand, estimated depth from the network does not have any missing values.  

The proposed model is outlined in Fig \ref{fig:overview}. Our initial goal in training is characterized by optimizing the loss function defined jointly for semantic categories and depth estimates: 
\begin{equation}
\label{eq:global_objective}
L= \lambda \times L_{sem} + L_{depth} 
\end{equation} 
In the above loss formulation $L_{depth}$ and $L_{segm}$ are optimized jointly using a shared representation in a multi-scale CNN model, yielding a per pixel response maps of predicted labels and depth estimates. In the final stage of optimization the interactions between these response maps will be incorporated in a joint CRF model and the whole model including the network parameters will be further refined to minimize the objective. 
The following two sections will introduce the network and described the details of the individual loss functions $L_{sem}$ and $L_{depth}$ and how they related to the network structure. In Section~\ref{sec:CRF} we  will elaborate on the CRF formulation.

\subsection{The model}
The network has two main modules; one for semantic segmentation and one for depth estimation. Both modules use the same set of features to accomplish their task. The shared part of a network, which is shown in blue in Fig \ref{fig:overview}, is a multi-scale network that extracts features from images.  It has been shown in the past  that multi-scale networks are effective in improving the performance of semantic segmentation, which is analogous to extraction of features at multiple scales~\cite{MurphyICLR15}\cite{EigenICCV15} in the traditional semantic segmentation approaches.
The convolutional feature maps in the last layers of each scale are shared between semantic segmentation and depth estimation branches which are shown in green and red in Fig \ref{fig:overview} respectively. The computed feature maps at different scales are upsampled and concatenated to form the comprehensive feature representation of the image. We chose to use the architecture of \cite{MurphyICLR15} because it produces denser output with stride of 8 using the atrous algorithm and  has smaller memory footprint. Feature sharing results in saving computational resources during test time and also boosts the performance as shown in Section~\ref{sec:results}. 

\subsection{Semantic Loss} 
For semantic segmentation module the network outputs a response map with the dimension of $C \times H \times W$ where $C$ is the number of semantic classes and $H$, $W$ are the height and width of input image. The semantic segmentation loss is accumulated per-pixel multinomial logistic loss which is equal to 
\begin{equation}
L_{sem} = -\sum_{i=1}^N \log\left(p(C_i^* | z_i ) \right)
\end{equation}
where $C^*_i$ is the ground truth label of pixel $i$,  $p(C_i | z_i)  = e^{z_i}/\sum_c e^{z_{i,c}}$ is the probability of estimating semantic category $C_i$ at pixel $i$, and $z_{i,c}$ is the output of the response map.  

\subsection{Depth Loss} 
In order to estimate the depth value, we divide the range of possible depth values to $N_d$  bins where each bin  has length $l$. For each bin $b$, the network predicts $p(b|x(i)) = e^{r_i}/\sum_b e^{r_{i,b}}$, the probability of having an object at the center of that bin and $r_{i,b}$ is the response of network at pixel $i$ and bin $b$.    The continuous depth value $d_i$ is the computed as: 
\begin{equation}
\label{eq:depth_value}
d_{i} = \sum_{b=1}^{N_d} b \times l \times p(b|x(i)) .
\end{equation}
One might think that it should be also possible to learn the discretized depth probability using multinomial logistic loss similar to semantic segmentation. In this case however the training diverges due to following reasons; (1) multinomial softmax loss is not suitable for depth because depth is a continuous quantity and it cannot properly account for the distance of  the estimated depth to the ground truth (it just indicates the estimated depth is incorrect);  (2) estimating absolute depth for each pixel is ambiguous due to absence of scene scale. Therefore we use scale-invariant loss function of \cite{EigenNIPS14} for depth estimation that tries to equalize the relative depth distance between any pair of points in the ground truth and the estimated depth values. Scale-invariant loss is computed as follows:
\begin{equation}
\label{eq:scale_invariant}
L_{depth} = \frac{1}{n^2} \sum_{i,j}{\left((\log(d_{i}) - \log(d_{j})) - (\log(d^*_{i}) - \log(d^*_{j}))\right)^2}
\end{equation}
The advantage of scale invariant loss is that it encourages to predict the correct relative depth of the objects with respect to each other rather than absolute depth values. Since we are exploiting depth discontinuities in the CRF, scale invariant loss is suitable for our setup.
\subsection{Conditional Random Field}
\label{sec:CRF}
\begin{figure}
 \begin{tabular}{c@{\hspace{1mm}}c@{\hspace{1mm}}c@{\hspace{1mm}}c@{\hspace{1mm}}c}
     
\includegraphics[width=0.2\linewidth]{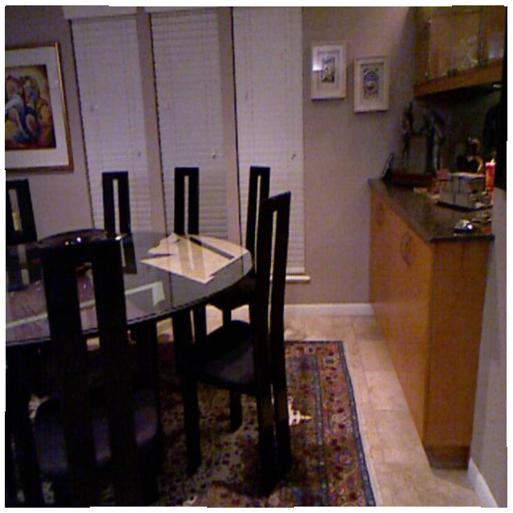}&
\includegraphics[width=0.2\linewidth]{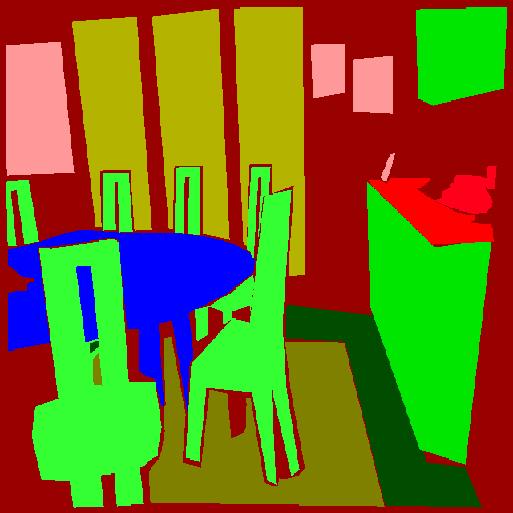}&
\includegraphics[width=0.2\linewidth]{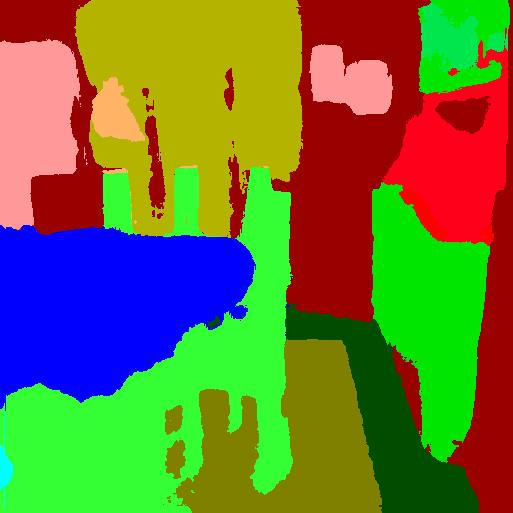}&
\includegraphics[width=0.2\linewidth]{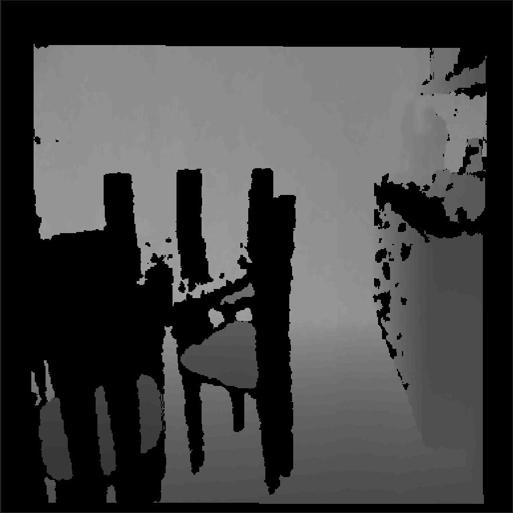}&
\includegraphics[width=0.2\linewidth]{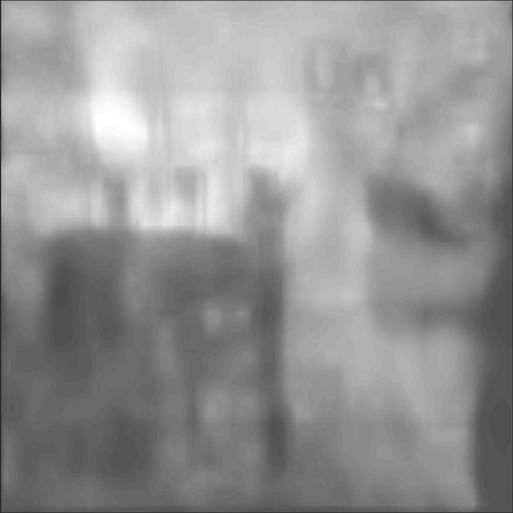}\\
\includegraphics[width=0.2\linewidth]{eccv/53_img}&
\includegraphics[width=0.2\linewidth]{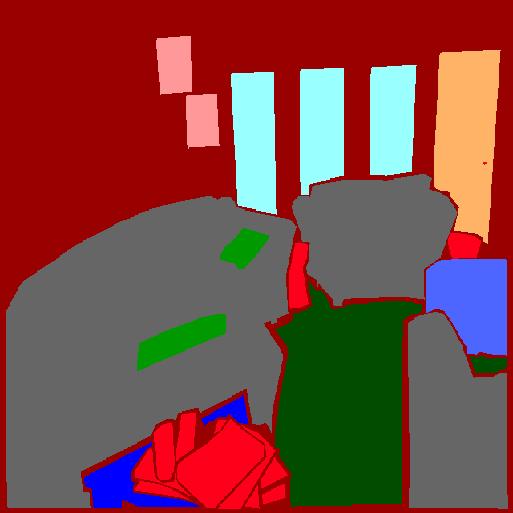}&
\includegraphics[width=0.2\linewidth]{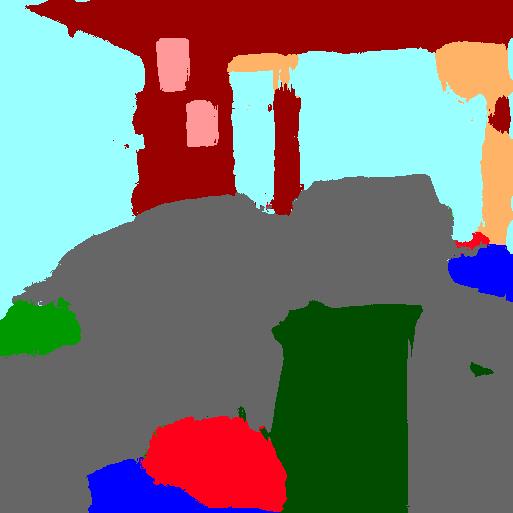}&
\includegraphics[width=0.2\linewidth]{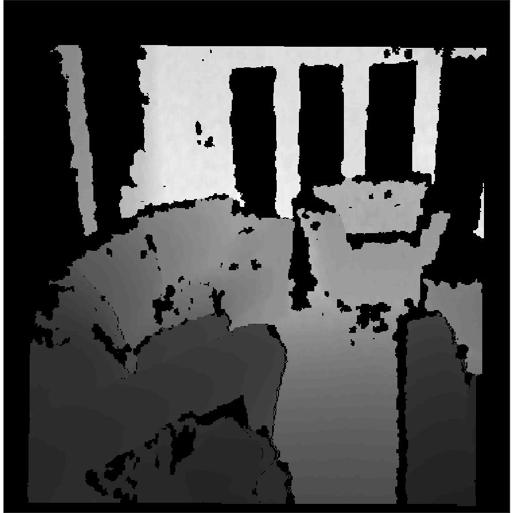}&
\includegraphics[width=0.2\linewidth]{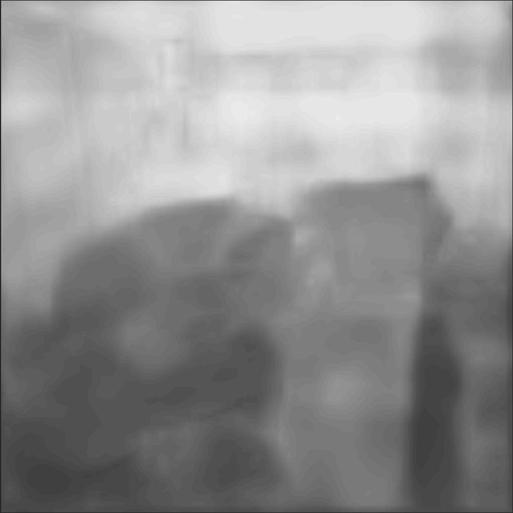}\\
\includegraphics[width=0.2\linewidth]{eccv/57_img}&
\includegraphics[width=0.2\linewidth]{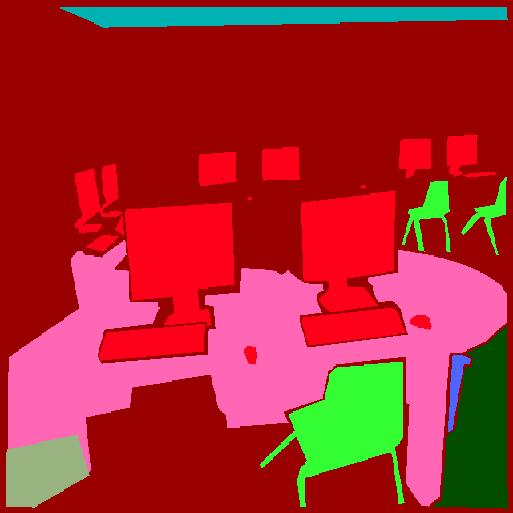}&
\includegraphics[width=0.2\linewidth]{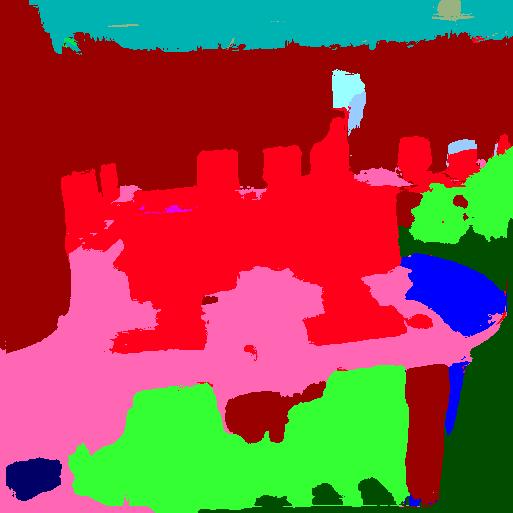}&
\includegraphics[width=0.2\linewidth]{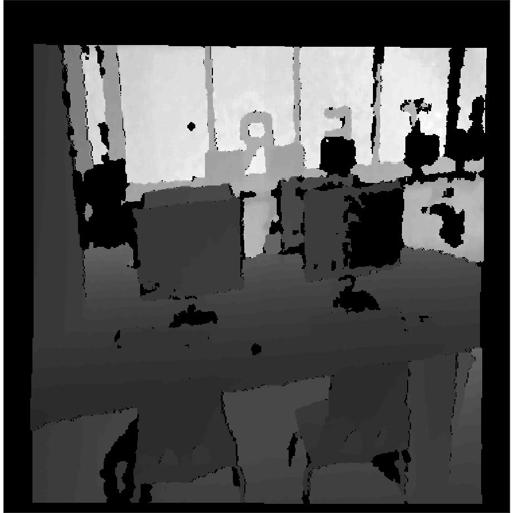}&
\includegraphics[width=0.2\linewidth]{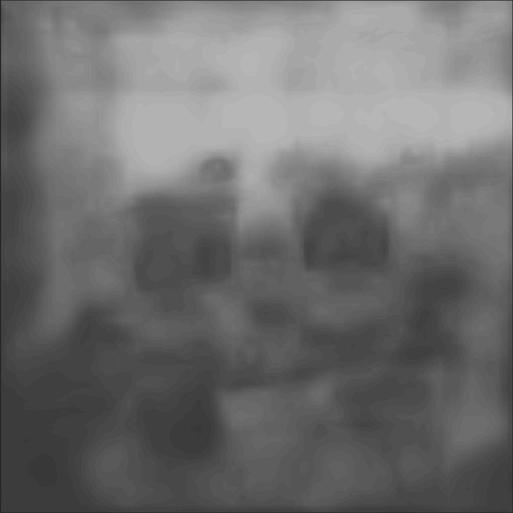}\\
\includegraphics[width=0.2\linewidth]{eccv/11_img}&
\includegraphics[width=0.2\linewidth]{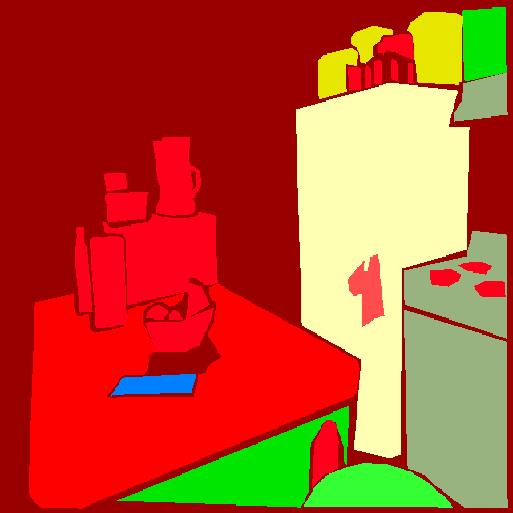}&
\includegraphics[width=0.2\linewidth]{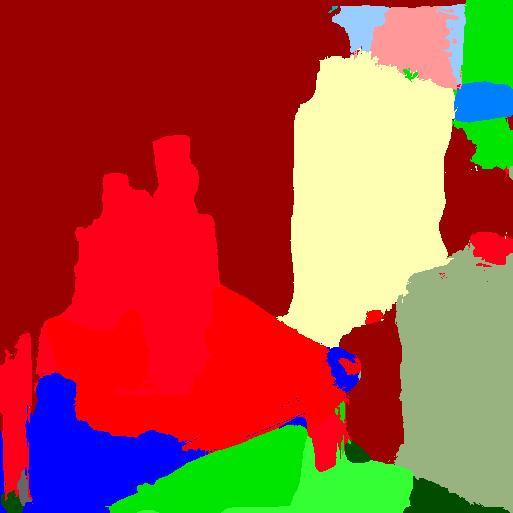}&
\includegraphics[width=0.2\linewidth]{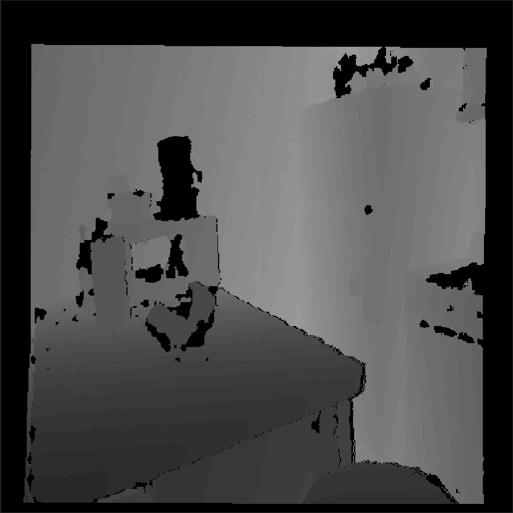}&
\includegraphics[width=0.2\linewidth]{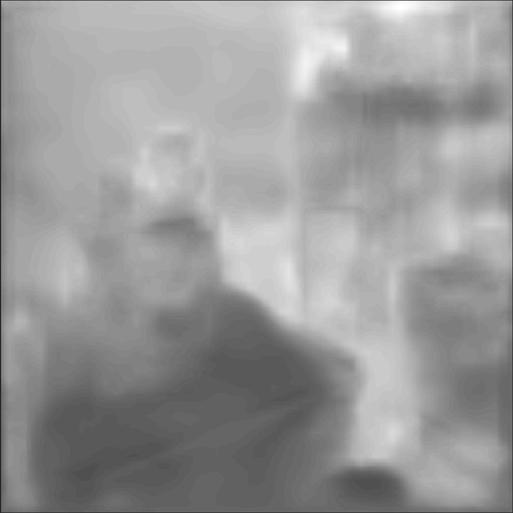}\\
\includegraphics[width=0.2\linewidth]{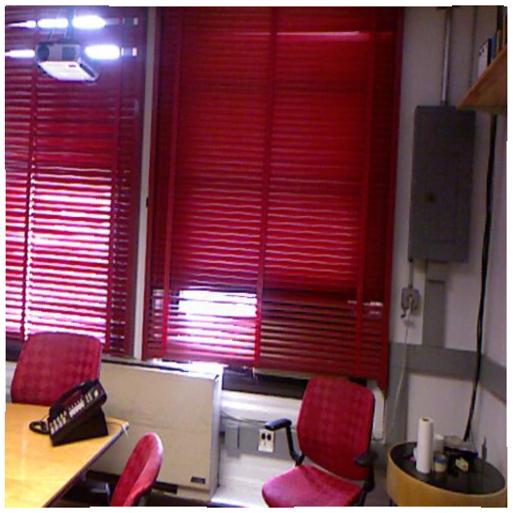}&
\includegraphics[width=0.2\linewidth]{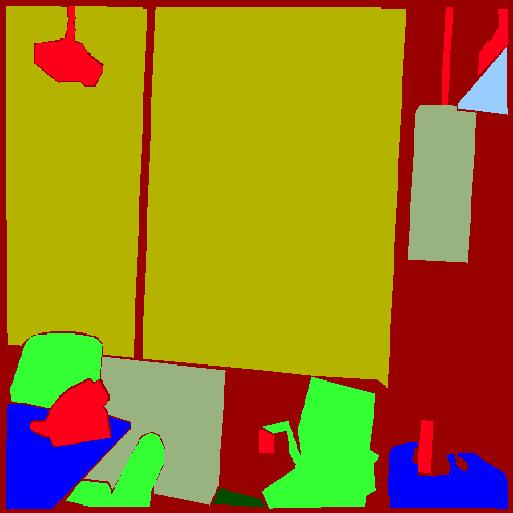}&
\includegraphics[width=0.2\linewidth]{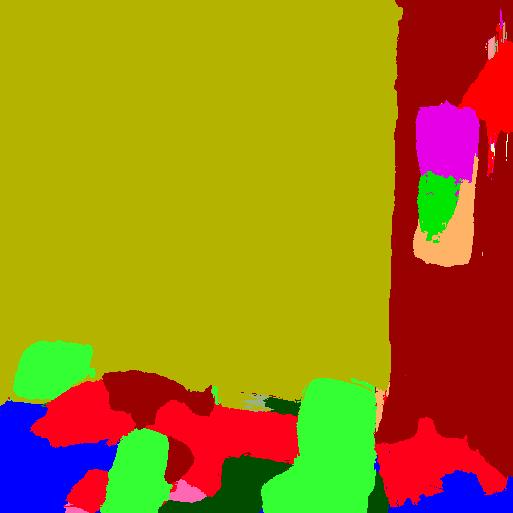}&
\includegraphics[width=0.2\linewidth]{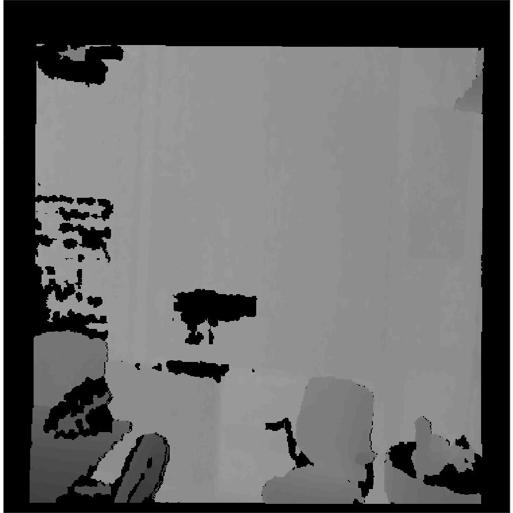}&
\includegraphics[width=0.2\linewidth]{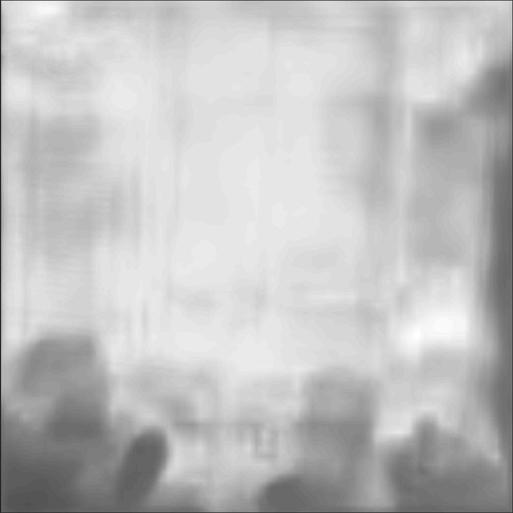}\\
\includegraphics[width=0.2\linewidth]{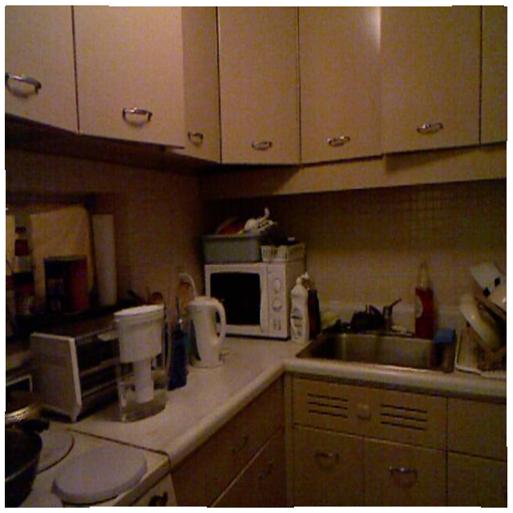}&
\includegraphics[width=0.2\linewidth]{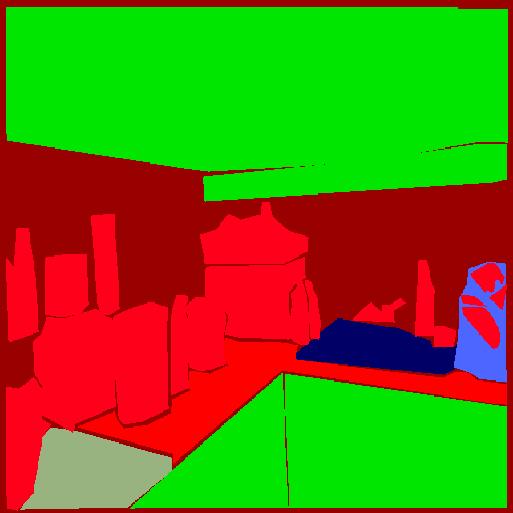}&
\includegraphics[width=0.2\linewidth]{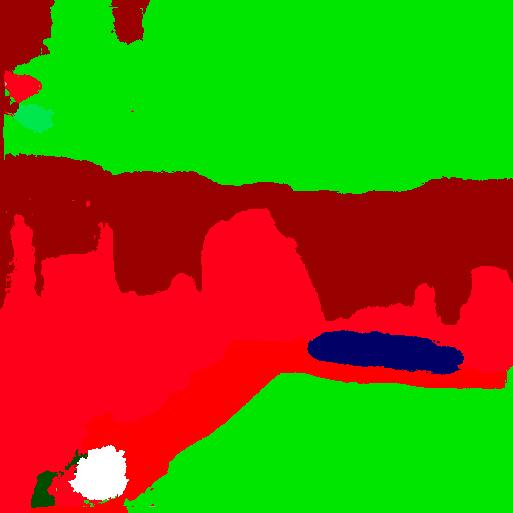}&
\includegraphics[width=0.2\linewidth]{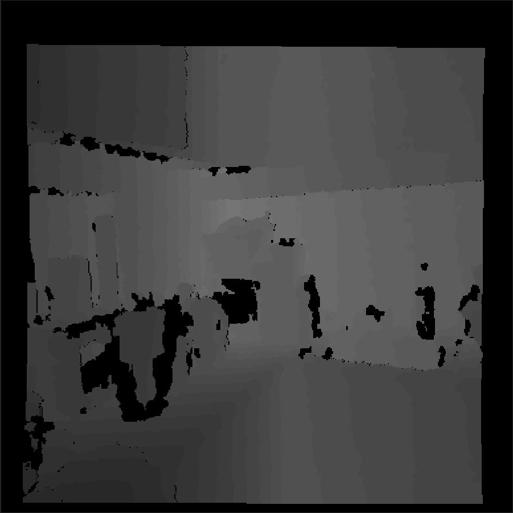}&
\includegraphics[width=0.2\linewidth]{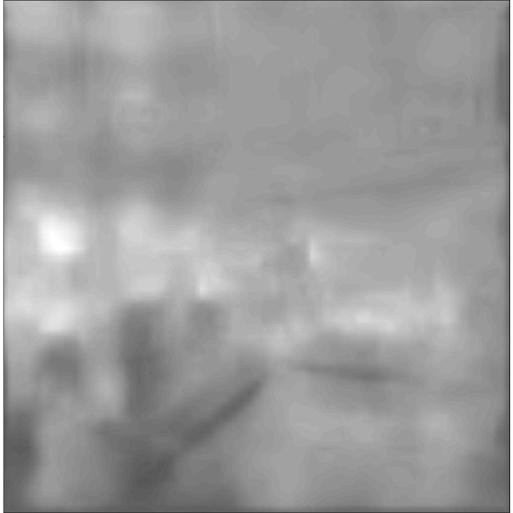}\\
(a) & (b) & (c) & (d) & (e)
\end{tabular}
\caption{Qualitative result of the proposed method. (a) is the input image (b) is the ground truth semantic segmentation (c) is the output of our semantic segmentation (d) is the raw depth and (e) is the estimated depths. Note that in the second rows our network detects the left window correctly whereas it is labeled as wall in the ground truth. The same situation happens in the third row where the left black chair is missing in the ground truth but our network detects it. The dark black region in the ground truth depth are the missing depths. However, we do not have to deal with missing depths in our output.}
\label{fig:res}
\end{figure}

\begin{figure}[t]
 \begin{tabular}{c@{\hspace{1mm}}c@{\hspace{1mm}}c@{\hspace{1mm}}c}
\includegraphics[width=0.23\linewidth]{eccv2/27_img}&
\includegraphics[width=0.23\linewidth]{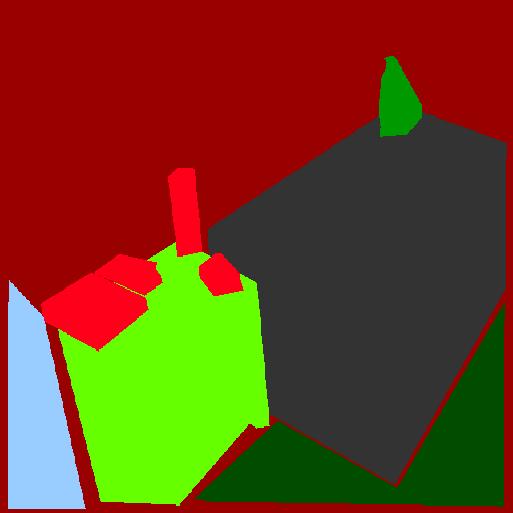}&
\includegraphics[width=0.23\linewidth]{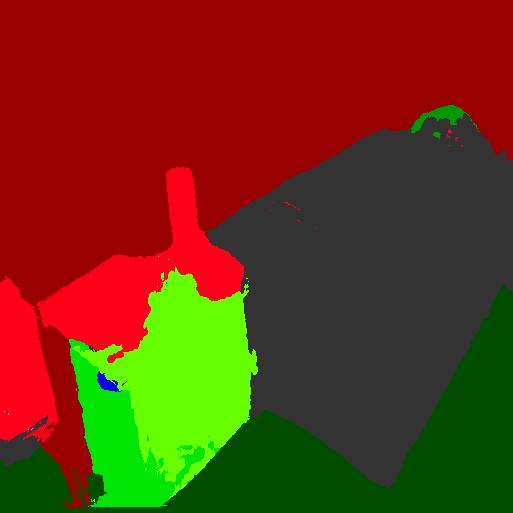}&
\includegraphics[width=0.23\linewidth]{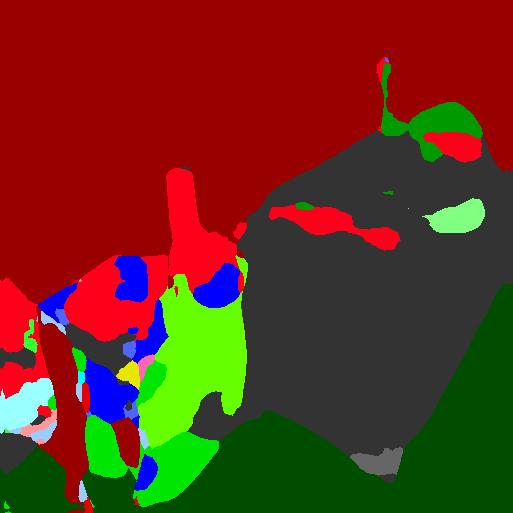}\\
\includegraphics[width=0.23\linewidth]{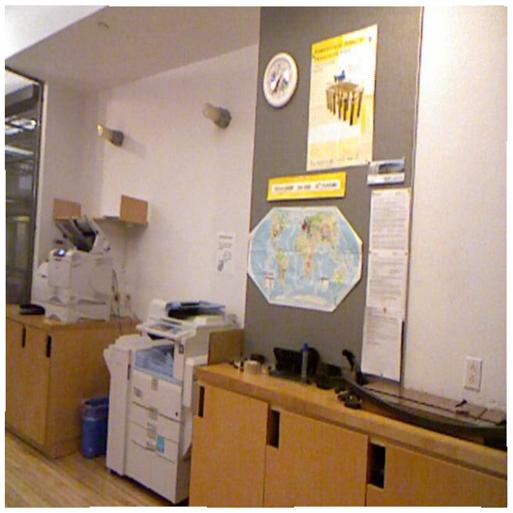}&
\includegraphics[width=0.23\linewidth]{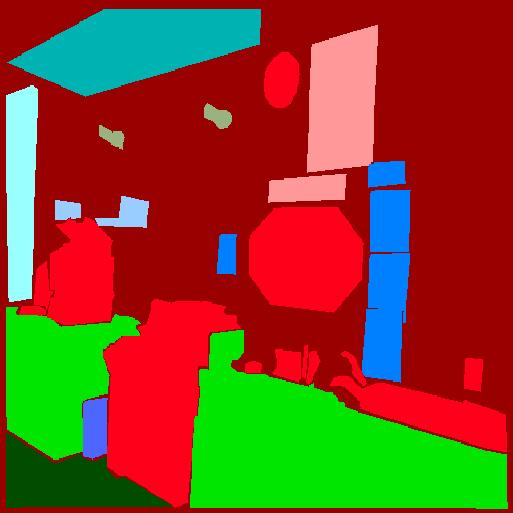}&
\includegraphics[width=0.23\linewidth]{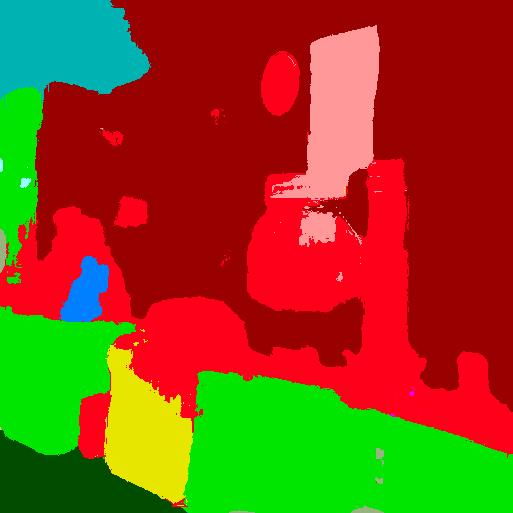}&
\includegraphics[width=0.23\linewidth]{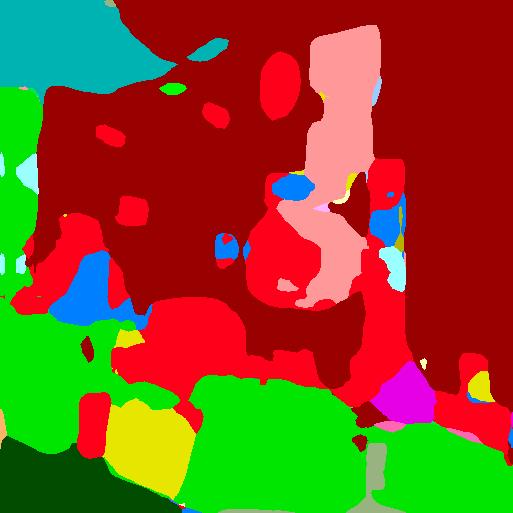}\\
\includegraphics[width=0.23\linewidth]{eccv2/56_img}&
\includegraphics[width=0.23\linewidth]{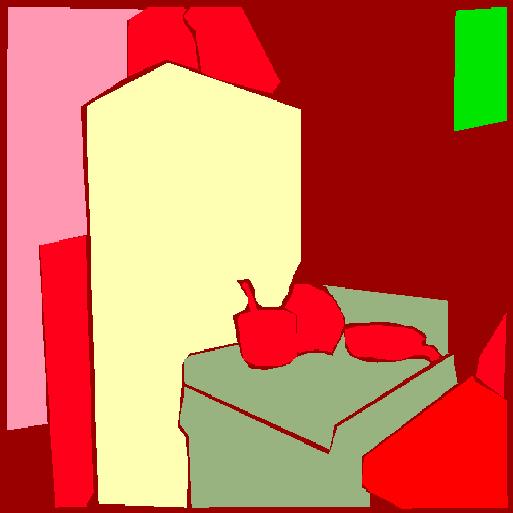}&
\includegraphics[width=0.23\linewidth]{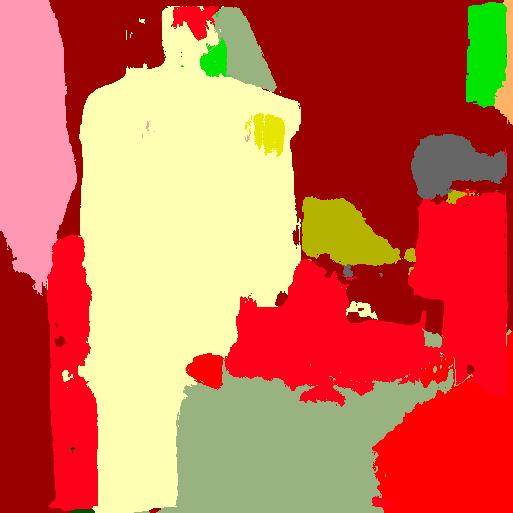}&
\includegraphics[width=0.23\linewidth]{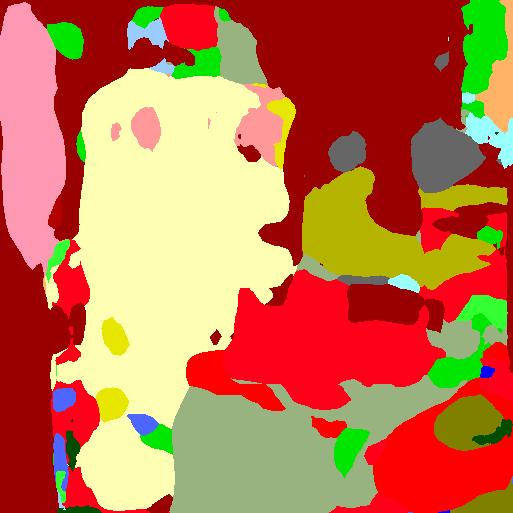}\\
(a) & (b) & (c) & (d)
\end{tabular}
\caption{Qualitative comparison of with and without CRF on semantic segmentation. (a) is input image, (b) is ground truth labeling, (c) is semantic segmentation with CRF, and (d) is the semantic unaries without CRF.}
\label{fig:CRF}
\end{figure}

As observed previously unary CNN based semantic segmentation results  showed that the response maps/labels are often not sufficiently well localized to achieve pixel accurate segmentation. This and the capability of capturing more general contextual relationships between semantic classes led to initial proposals for incorporating CRF's. 
Using these observations, we integrate the depth and semantic label predictions in the CRF framework. 
The unary potentials are computed from semantic output of the multi-scale network and pairwise terms are Gaussian potentials based on depth discontinuities, difference in RGB values of pixels and the compatibility between semantic labels. Let $N$ be the number of pixels and $X=\{x_1,x_2,...,x_N\}$ be the label assignment and $x_i \in \{1,...,C\}$. The features that we are using for each pixel $i$ are denoted by $f_i=\{p_i, I_i, d_i\}$ where $p_i$ is the spatial location of pixel $i$, $I_i$ is the RGB value of pixel $i$, and $d_i$ is the estimated depth at pixel $i$. The energy function for the fully connected CRF is follows:
\begin{equation}
\label{eq:crf_energy}
E(x,f) = \sum_i \psi_{u}(x_i) + \sum_{i,j} \psi_p(x_i, f_i, x_j, f_j)
\end{equation}
where unary potentials $\psi_u(x_i)$ come from the multi-scale network (the left big green rectangle in Fig \ref{fig:overview}) and the pairwise potentials have the form 
\begin{equation}
\psi_p(x_i, f_i, x_j, f_j) = \mu(x_i,x_j) k(f_i, f_j)
\end{equation}
where $\mu(x_i,x_j)$ represents the compatibility function between semantic label assignments of pixel $i$ and $j$. 
Gaussian kernel $k(f_i, f_j)$ adjusts the evidence that should be propagated between $x_i$ and $x_j$ based on the spatial distance, RGB distance, and depth distance between pairs of pixels . $k(f_i, f_j)$ consists of three different weights $\{w^{(i)}|i \in \{1,2,3\}\}$ where each $w^{i}$ has $C \times C$ parameters that are being learned for all the pairs of semantic categories. Gaussian kernels also have hyper-parameters $\theta_{(.)}$ that control the tolerance with respect to difference in depth values, RGB pixel values and spatial location of pairs of pixels. $k(f_i,f_j)$ is computed using the following equation:
\begin{equation}
\label{eq:gaussian_kernel}
\begin{split}
k(f_i,f_j) & =  w^{(1)} exp\left(\frac{|p_i - p_j|^2}{2\theta_{\alpha}^2}+\frac{|I_i - I_j|^2} {2\theta_{\beta}^2} \right) \\
 & + w^{(2)} exp\left(\frac{|p_i - p_j|^2}{2\theta_{\gamma}^2}+\frac{|d_i - d_j|^2}{2\theta_{\zeta}^2} \right) \\
  & + w^{(3)} exp\left( \frac{|p_i - p_j|^2}{2\theta_{\tau}^2} \right)
\end{split}
\end{equation}
The inference in the CRF is done using mean-field approximation similar to \cite{TorICCV15}. In the CRF training stage both the compatibility terms, the kernel weights and unary potentials are learned in a single optimization procedure. The derivatives are back propagated through the network further refining the shared feature representation captured by network weights.  Note that the CRF only adjusts its weights and back-propagates the error only to the semantic unaries and shared layers through the semantic module. Estimated depths are only taken as extra input modality in the CRF.  However since both $L_{sem}$ and $L_{depth}$ is still being optimized the depth convolution layers will be adjusted to keep the output depth values valid.  In the following section, we present additional details of multi-stage optimization and the scrutinize the effects of different components of the loss function on the overall performance. Qualitative results are shown in Fig. \ref{fig:res} and Fig. \ref{fig:CRF}. Fig \ref{fig:res} shows the qualitative results of joint depth estimation and semantic segmentation. It is worth noting that  there are some cases where our network detects a category correctly but that category is labeled incorrectly in the dataset. Two examples of such situations are the the left window and the leftmost chair in front of the desk in the second and third rows of Fig. \ref{fig:res}. Fig \ref{fig:CRF} shows qualitative effect of the CRF module on the output of semantic segmentation.

\section{Experiments}
\label{sec:results}
 
Before we proceed with details on the performance evaluation, we present in more detail the parameters of the network. The shared part of a network, which is shown in blue in Fig. \ref{fig:overview} is a multi-scale network that extracts features from the images. The details about the parameters of the layers are found in Table~\ref{table:net_params}.
The first dimension is the number of channels for the output and the rest is the kernel size of that layer. The network has 5 different branches each either takes and image or one of the earlier layers as input and computes more higher-level features. The input resolution is 513 $\times$ 513 and at the end of each branch the computed features for semantics and depth are resized so to the dimension of  the image size.
\begin{table*}
\caption{Details of multi-scale network for computing depth and semantic unaries. Dimensions of each layer shown in the number of output channels and the kernel size.}
\label{tab:net_details}
\scalebox{0.8}{
\begin{tabular}{|c|c|cccccccc|}
\hline
Branch & Input &  & & & & & & & \\
\hline
Branch1& RGB & \begin{tabular}{c} conv1-1 \\ 64x3x3 \end{tabular} & \begin{tabular}
{c} conv1-2 \\ 64x3x3 \end{tabular} & \begin{tabular}{c} conv1-seg \\ 40x1x1 
\end{tabular} & \begin{tabular}{c} conv1-depth \\ 50x1x1 \end{tabular} &  
 & & & \\
\hline
Branch2& RGB & \begin{tabular}{c} conv2-1 \\ 64x3x3 \end{tabular} & \begin{tabular}
{c} conv2-2 \\ 64x3x3 \end{tabular} & \begin{tabular}
{c} pool2 \\ 64x3x3 \end{tabular} & \begin{tabular}{c} conv2-3 \\ 128x3x3 \end{tabular} & \begin{tabular}{c} conv2-seg \\ 40x1x1 
\end{tabular} & \begin{tabular}{c} conv2-depth \\ 50x1x1 \end{tabular} & \begin{tabular}{c}
upsample \\ x2
\end{tabular} & \\
\hline
Branch3& pool2 & \begin{tabular}{c} conv3-1 \\ 128x3x3 \end{tabular} & \begin{tabular}
{c} conv3-2 \\ 128x3x3 \end{tabular} & \begin{tabular}
{c} pool3 \\ 128x3x3 \end{tabular} & \begin{tabular}{c} conv3-3 \\ 128x3x3 \end{tabular} & \begin{tabular}{c} conv3-4 \\ 128x3x3 \end{tabular} & \begin{tabular}{c} conv3-seg \\ 40x1x1 
\end{tabular} & \begin{tabular}{c} conv3-depth \\ 50x1x1 \end{tabular} & \begin{tabular}{c}
upsample \\ x4
\end{tabular} \\
\hline
Branch4& pool3 & \begin{tabular}{c} conv4-1 \\ 256x3x3 \end{tabular} & \begin{tabular}
{c} conv4-2 \\ 256x3x3 \end{tabular} & \begin{tabular}
{c} pool4 \\ 256x3x3 \end{tabular} & \begin{tabular}{c} conv4-3 \\ 128x3x3 \end{tabular} & \begin{tabular}{c} conv4-4 \\ 128x3x3 \end{tabular} & \begin{tabular}{c} conv4-seg \\ 40x1x1 
\end{tabular} & \begin{tabular}{c} conv4-depth \\ 50x1x1 \end{tabular} & \begin{tabular}{c}
upsample \\ x4
\end{tabular}\\
\hline
Branch5& pool4 & \begin{tabular}{c} conv5-1 \\ 512x3x3 \end{tabular} & \begin{tabular}
{c} conv5-2 \\ 512x3x3 \end{tabular} & \begin{tabular}
{c} pool5 \\ 512x3x3 \end{tabular} & \begin{tabular}{c} conv5-3 \\ 1024x3x3 \end{tabular} & \begin{tabular}{c} conv5-4 \\ 1024x1x1 \end{tabular} & \begin{tabular}{c} conv5-seg \\ 40x1x1 
\end{tabular} & \begin{tabular}{c} conv5-depth \\ 50x1x1 \end{tabular} &\begin{tabular}{c}
upsample \\ x8
\end{tabular} \\
\hline
\end{tabular}}
\label{table:net_params}
\end{table*}

\subsection{Training Details}
Training is done at multiple stages. The training objective function for stage 1 is only $L_{sem}$ and for the rest of the stages, Eq. \ref{eq:global_objective} is optimized for the training. In the first stage of training, the network is trained for 160K iterations with learning rate of $1e$-10, weight decay of $0.0005$ and momentum of $0.99$ for semantic segmentation. The network weights of stage 1 are initialized from the model of \cite{MurphyICLR15} which is pre-trained on MS-COCO dataset and fine-tuned on Pascal-VOC dataset.

In the second stage, the depth layers (shown in red in Fig \ref{fig:overview})  are added to the network that is already 
trained on semantic segmentation. The network is initialized with the previous stage weights and is trained using combined semantic segmentation and depth estimation loss for 10K iterations. The scale of semantic and depth loss are different. Therefore, the effect of these loss functions should be balanced through the weight $\lambda$ in Eq. \ref{eq:global_objective}. The $\lambda$ was set to $1e$-6 to balance semantic loss and depth loss objectives. We also tried training with $L_{depth}$ and $L_{sem}$ together instead of two stages of training. We observed that with the joint training, the value of objective function dropped much quicker but plateaued at the end. The two-stage training resulted in a slightly better model.

 In the third stage, the fully connected CRF was added to the network fine-tunning the network jointly to learn the CRF weights. We used learning rate of $1e$-13 for the CRF weights and learning rate of $1e$-16 for the rest of network and ran the training for 10K iterations. In order to train the CRF,  $w^{(1)}$ is initialized to 7, $w^{(2)}$ to 4, and $w^{(3)}$ is initialized with 3.  The remaining parameters  $\theta_{\alpha}$ to 160, $\theta_{\beta}$ to 3, $\theta_{\gamma}$ to 50, $\theta_{\zeta}$ to 0.2, and $\theta_{\tau}$ to 3. All the initialization and hyper-parameters are found by cross validation on a random subset of 100 images from training set.

We trained and evaluated the model on NYUDepth v2 dataset \cite{SilbermanECCV12} using the standard train/test split. The training set contains 795 images and the test set contains 654 images. For training the dataset is augmented  by cropping, and mirroring. For each image, we generated 4 different crops and scale the depth accordingly. In addition, the original image and its mirrored image were also included in the training set, yielding 
4770 images from original training set. The data augmentation procedure was done offline and the data was shuffled randomly once before the training. The following sections contains the evaluation of our method on depth estimation and semantic segmentation.
\begin{figure*}[]
\centering
\begin{tabular}{ccc}
\includegraphics[width=0.3\textwidth]{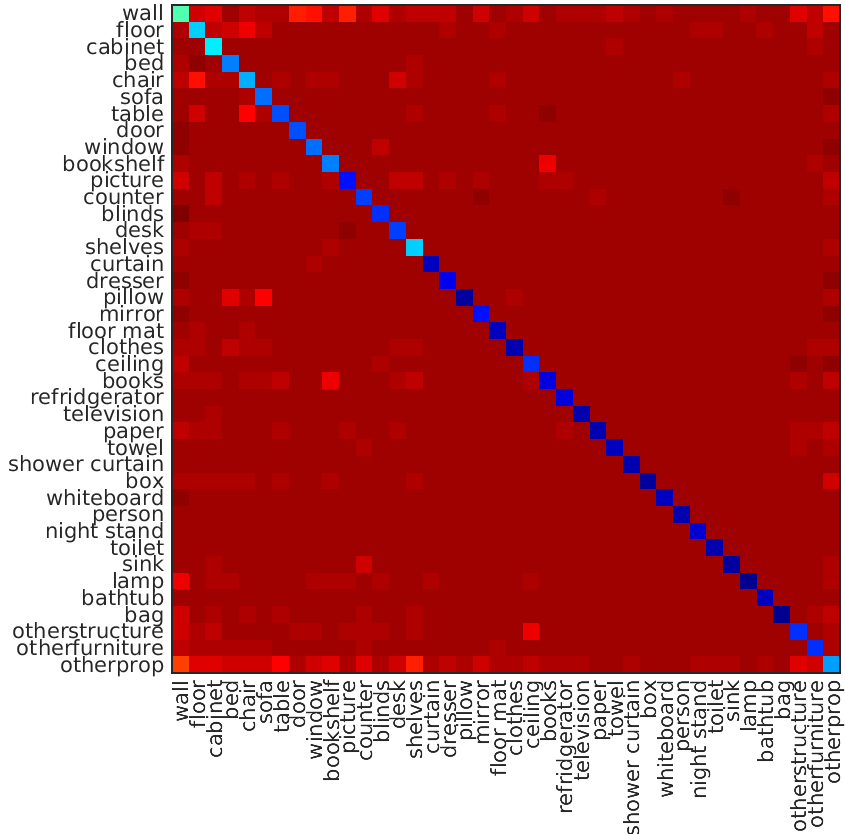}&
\includegraphics[width=0.3\textwidth]{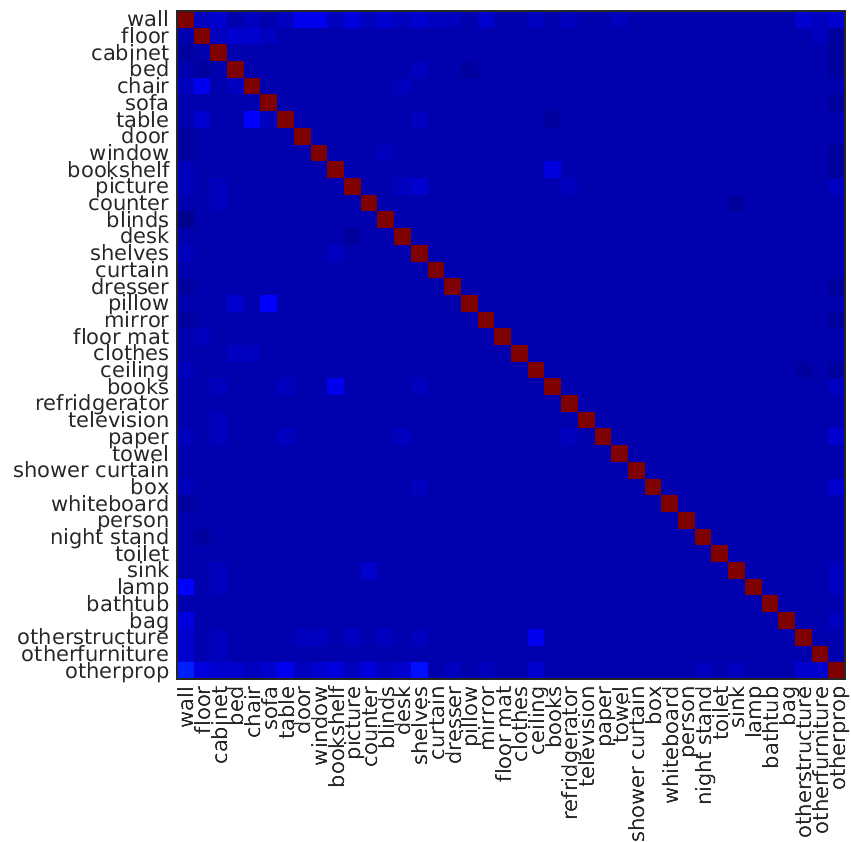}&
\includegraphics[width=0.3\textwidth]{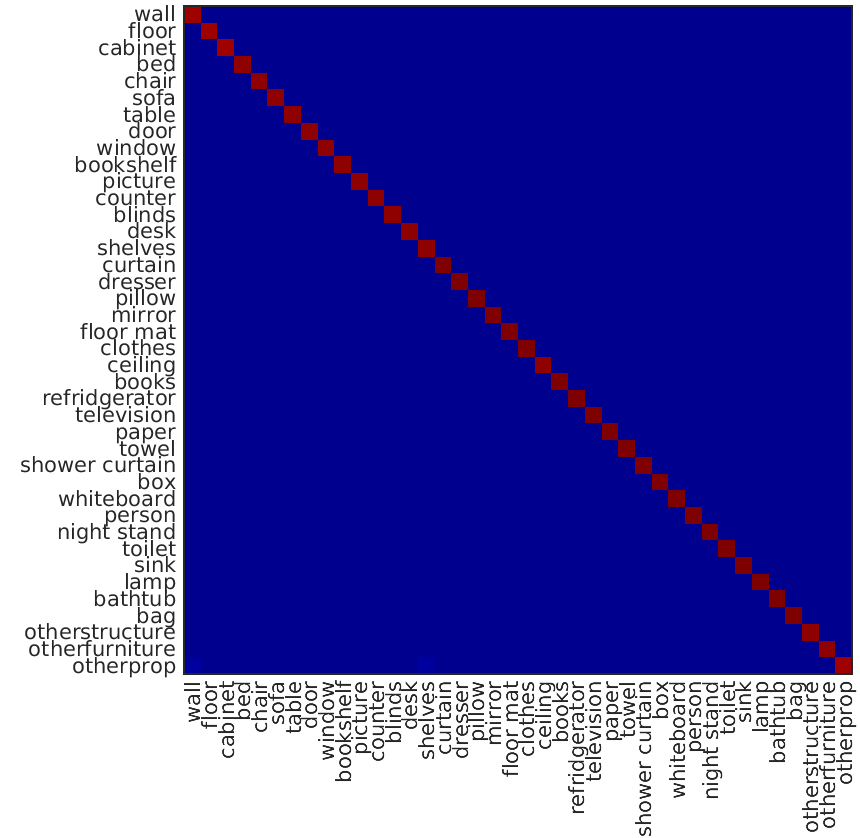}\\
\end{tabular}
\caption{Visualization of learned weights in CRF. Left: compatibility function $\mu(.,.)$ between classes, middle: learned weights $w^{(2)}$ for depth for all pairs of semantic classes, right: learned weights $w^{(1)}$ for difference in RGB value of each pixel for all pairs of semantic classes (best viewed electronically).}
\label{fig:weights}
\end{figure*} 
\subsection{Depth Estimation}
For depth estimation, we use $N_d=50$ bins with the length of $l=0.14$m in the network. After applying softmax and using Eq~\ref{eq:depth_value}, depth value is estimated. We modified the ground truth depth values in the training to make the joint problem of semantic segmentation and depth estimation less complex and also reduce the uncertainty from the depth readings. Note that the ground truth depth values of evaluation set remain intact. The ground truth depth values are clipped at $7$m because the quality of raw depth values from RGB-D decreases with the depth and the farther sensor readings are not reliable. We also rounded the depth value to the closest multiplier of $l$. We only used the valid depth values for training.  Quantitative evaluation of our method is shown in Table \ref{tab:depth}. Our method outperforms only on the scale invariant loss which is sensitive to relative order of the entities with respect to each other. Given that the network is trained under multiple objective functions and learning relative ordering of the object is enough for reasoning in semantic space, it is reasonable that the network performs well only on scale invariant loss.
\begin{enumerate}
\item Percentage of Depth $d_i$ where the ratio of estimated and ground truth depth is less than a threshold. i.e. $max(\frac{d_i}{d^*_i}, \frac{d_i}{d^*_i}) = \delta < threshold$.
\item Absolute Relative Difference: $\frac{1}{T} \sum |d_i - d^*_i| / d^*_i$
\item Squared Relative Difference: $\frac{1}{T} \sum |d_i - d^*_i|^2 / d^*_i$
\item RMSE (linear): $\sqrt{\frac{1}{|T|} \sum ||d_i - d^*_i||^2}$
\item RMSE (log): $\sqrt{\frac{1}{|T|} \sum ||log (d_i) - log (d^*_i)||^2}$
\item RMSE (log scale-invariant): equals to RMSE (log) after equalizing the mean estimated depth and ground truth depth.
\end{enumerate}
where $d$ and $d^*$ are the estimated depth and ground truth depth respectively.
Note that our RMSE error for scale invariant is significantly better and it quantitatively shows that our method is much better in finding depth discontinuities because scale invariant error, as the name implies, emphasizes on the relative depth not the absolute value of depth.
\begin{table*}[t]
\centering
\caption{Quantitative Evaluation of Depth Estimation}
\label{tab:depth}
\begin{tabular}{|c|c c c|c|}
\hline
 & Eigen et al.\cite{EigenICCV15} & Liu et al \cite{LiuCVPR15} & Ours &\\
\hline
threshold $\delta < 1.25$ & {\bf 0.769} & 0.614 & 0.568 & higher \\
threshold $\delta < 1.25^2$ & {\bf 0.950} & 0.883 & 0.856 & is \\
threshold $\delta < 1.25^3$ & {\bf 0.988} & { 0.971} & 0.956 & better \\
\hline
abs relative distance & {\bf 0.158} & 0.230 & {0.200} & \\
sqr relative distance & {\bf 0.121} & - & 0.301 & lower\\
RMSE (linear) & {\bf 0.641} & 0.824 & {0.816} & is \\
RMSE (log) & {\bf 0.214} & - & 0.314 & better \\
RMSE (log. scale invariant) & 0.171 & - & {\bf 0.061}& \\
\hline
\end{tabular}
\end{table*}
\subsection{Semantic Segmentation}
Semantic segmentation was evaluated on 40 semantic labels of NYUDepth V2 dataset using the mean Intersection over Union (IoU) which is the average Jaccard score among all the classes. Mean accuracy is the average pixel accuracy among all classes and pixel accuracy is the total accuracy of pixels regardless of the category. 
\begin{table*}[]
\centering
\caption{Quantitative Evaluation of Semantic Segmentation on 40 Categories of NYUDepth v2} 
\label{tab:seg_res}
\begin{tabular}{|c | c | c  c c|}
\hline
Method & Input Type & Mean IoU & Mean Accuracy & Pixel Accuracy\\
\hline
Deng at all \cite{SiniaICCV15} & RGBD & N/A & 31.5 & 63.8 \\
FCN\cite{LongCVPR15} & RGB  & 29.2 & 42.2 & 60.0 \\
FCN + Depth \cite{LongCVPR15} & RGBD & 34.0 & 46.1 & 65.4 \\ 
Eigen and Fergus \cite{EigenICCV15} & RGB & 34.1 & 45.1 & 65.6 \\
\hline
Ours-Unary-Sem & RGB & 36.0 & 49.1 & 66.0 \\
Ours-Unary-Sem+Depth & RGB & 36.5 & 49.2 & 66.6 \\
Ours-Sem-CRF & RGB & 38.4 & 51.2 & 68.0 \\
Ours-Sem-CRF+ & RGB & {\bf 39.2} & {\bf 52.3} & {\bf 68.6}\\
\hline
\end{tabular}
\label{tab:results}
\end{table*}
As shown in Table \ref{tab:seg_res}, our method outperforms the recent methods. Our-Unary-Sem is the performance of the network when only trained on semantic segmentation without depth (Training Stage 1). Ours-Unary-Sem+Depth is the network with semantic and depth without depth (Training Stage 2). Ours-Sem-CRF is the result of having both semantic and depth unaries but the CRF uses only RGB pixel values and semantic unaries as input. Our-Sem-CRF+ is including all the modules and CRF takes both the estimated depth and RGB pixel values as input. Overall, estimating the depth in addition to semantic segmentation improves the mean IoU over 40 classes by 1.3\%. Similar observation is reported in \cite{WangCVPR15}, however our method is 10x faster and everything is trained end-to-end.

In order to further investigate how the CRF uses the depth information, $w^{(1)}$ and $w^{(2)}$ are visualized in Fig \ref{fig:weights}.  Note that the difference in RGB values is not informative as the weights for differences in depth values between pixels.  One interesting observation is that $w^{(2)}$ is large for pairs of classes where the depth discontinuity helps. Some of the examples of such pairs are pillow vs couch, bookshelf vs book, and sink vs counter.

\section{Conclusions}
We showed how to do semantic segmentation and depth estimation jointly using the same network which is trained in stages and then fine tuned using a single loss function.  
The proposed model and the training procedure produces comparable depth estimates and superior semantic segmentation comparing to state-of-the-art methods.  Moreover, we showed that coupling CRF with the deep network further improves the performance and enables us to  exploit the estimated depth to discriminate between some of the semantic categories. Our results show that depth estimation and semantic segmentation can share the underlying feature representations and can help to improve the final performance. 

\section{Acknowledgements}
Supported by the Intelligence Advanced Research Projects Activity (IARPA) via Air Force Research Laboratory contract FA8650-12-C-7212. The U.S. Government is authorized to reproduce and distribute reprints for Governmental purposes notwithstanding any copyright annotation thereon. Disclaimer: The views and conclusions contained herein are those of the authors and should not be interpreted as necessarily representing the official policies or endorsements, either expressed or implied, of IARPA, AFRL, or the U.S. Government. We also acknowledge support from NSF NRI grant 1527208. Some of the experiments were run on ARGO, a research computing cluster provided by the Office of Research Computing at George Mason University.

{\small
\bibliographystyle{ieee}
\bibliography{egpaper_for_review}
}

\end{document}